\documentclass[letterpaper]{article} 
\usepackage[]{aaai23}  
\usepackage{times}  
\usepackage{helvet}  
\usepackage{courier}  
\usepackage[hyphens]{url}  
\usepackage{graphicx} 
\usepackage{natbib}  
\usepackage{caption} 
\usepackage{algorithm}
\usepackage{algorithmic}

\usepackage{newfloat}
\usepackage{listings}
\DeclareCaptionStyle{ruled}{labelfont=normalfont,labelsep=colon,strut=off} 
\floatstyle{ruled}
\newfloat{listing}{tb}{lst}{}
\floatname{listing}{Listing}
\title{ReGrAt: Regularization in Graphs using Attention to handle class imbalance}
\author{
    Neeraja Kirtane\textsuperscript{\rm 1}, Jeshuren Chelladurai\textsuperscript{\rm 1}, Balaraman Ravindran\textsuperscript{\rm 1}, Ashish Tendulkar\textsuperscript{\rm 2}
}
\affiliations{
    \textsuperscript{\rm 1}Robert Bosch Centre for Data Science and Artificial Intelligence (RBC-DSAI)\\

   IIT Madras, India\\
   kirtane.neeraja@gmail.com
   
   \textsuperscript{\rm 2} Google, India
}

\usepackage{bibentry}

\begin{document}

\maketitle

\begin{abstract}
Node classification is an important task to solve in graph-based learning. Even though a lot of work has been done in this field, imbalance is neglected. Real-world data is not perfect, and is imbalanced in representations most of the times.  Apart from text and images, data can be represented using graphs, and thus addressing the imbalance in graphs has become of paramount importance. In the context of node classification, one class has less examples than others. Changing data composition is a popular way to address the imbalance in node classification. This is done by resampling the data to balance the dataset. However, that can sometimes lead to loss of information or add noise in the dataset. Therefore, in this work, we implicitly solve the problem by changing the model loss. Specifically, we study how attention networks can help tackle imbalance. Moreover, we observe that using a regularizer to assign larger weights to minority nodes helps to mitigate this imbalance. We achieve State of the Art results than the existing methods on several standard citation benchmark datasets.
\end{abstract}

\section{Introduction}
Many instances of real-world data can be represented in the form of graphs. 
One important task in graph learning is semi-supervised node classification \cite{yang2016revisiting}, where only a small amount of nodes are labeled. \citet{kipf2016semi} solves this problem using a basic GCN architecture. However, the architectures for node classification do not address the imbalance in the data and only perform well in a balanced setting. We say that data is imbalanced in the context of node classification when the number of nodes of one class is lesser than others. We call that class the minority class and the rest majority classes.  \par
One real-life example of imbalance in graphs is the problem of fraud detection, where the members committing fraud are in the minority compared to the number of users not involved in any fraud. Here, the nodes in the graph would be the actual users or bots, and the edges would represent the interaction between these nodes. The number of bots will be very less compared to the actual users. The problem becomes harder when the setup is semi-supervised as the size of the labeled training set is very low. The trained model works well on the majority class and poorly on the minority class.\par
There are three broad categories to tackle imbalance in machine learning: data-level, model-level, and hybrid approaches.
The data-level methods resample the data to make the dataset more balanced. However, that cannot be done every time as it leads to data loss, and sometimes, the data can lose its physical significance. For example, if we resample a protein structure by adding or deleting nodes, it would lose its structural identity. Therefore, is a need to implicitly handle the imbalance by making changes to the model than the data. The model-level approaches modify the loss function in such a way as to focus more on the minority nodes. Hybrid approaches combine both methods and create an ensemble of the models in the end.\par
In this work, we propose a model-based approach where we use Graph Attention networks with a regularizer to handle the imbalance. The regularizer is used by making the attention weights focus more on the minority nodes than the majority nodes. This ensures that the weights for minority nodes are higher than before, and hence they would be classified correctly. 
 To the best of our knowledge, this is the first work that takes care of imbalance in graph networks while implicitly keeping the data intact.
 Our major contributions to this paper are as follows:
\begin{itemize}
    \item Implicitly solve the problem of class imbalance by changing the model loss.
    \item Use attention mechanism to handle the class imbalance.
    \item Our results outperform the existing results that tackle imbalance handling for citation network datasets.

\end{itemize}
\section{Related work}
\subsection{Class imbalance}
Handling class imbalance is a long-studied problem in the context of Machine Learning, and much work has been done previously on it. As mentioned earlier, the methods are broadly classified into three types: data-level, model-level, and hybrid approaches.\par
Data level methods usually resample the data in a way to make the distribution more balanced than before. SMOTE \cite{chawla2002smote} is a widely used state-of-the-art method that handles data imbalance by adding synthetic data to make data balanced. In SMOTE, artificial samples are produced by interpolating between existing minority samples and their nearest minority neighbors. SMOTE has been tried in the context of graphs to tackle imbalance where new nodes are created from minority classes. \cite{zhao2021graphsmote}. An edge generator is used to find out the links between new nodes and the existing ones. An embedding space is constructed to encode the similarity among the nodes and generate new samples.
In addition, an edge generator is trained simultaneously to model the relational information and provide it for those new samples.\par
Model-level methods used till now use a cost-sensitive approach which increases the priority of minority classes. The loss function is reweighted to achieve this. Hybrid approaches train multiple models using the methods mentioned above and then create an Ensemble of the models to get the final results \cite{ensemble}.
\subsection{Graph Neural Networks}
Graph Neural Networks (GNNs) have received much attention and developing rapidly because of the need to work on non-linear structured data. Graph Convolutional Networks (GCNs) \cite{kipf2016semi} perform similar operations like CNNs. The node in the model learns features by 
inspecting the neighboring nodes. The major difference between CNNs and GNNs is that 
CNNs are specially built to operate on structured data, whereas GNNs are the generalized 
version of CNNs where the numbers of node connections vary and the nodes are unordered.
CNNs are used in Euclidean data and GNNs are used in non-Euclidean data. Current GNNs follow a message-passing framework, which is composed of pattern extraction and interaction modeling within each layer. Graph Attention Networks \cite{velivckovic2017graph} overcome the shortcoming of GCNs by having dynamic weights instead of having static ones. This is done by using masked self-attention layers. 

\section{Proposed Method}
\begin{table*}
\centering
\begin{tabular}{|l|lll|lll|}
\hline
\textbf{Methods}     & \multicolumn{3}{c|}{\textbf{Cora}}                                                     & \multicolumn{3}{c|}{\textbf{CiteSeer}}                                                 \\ \hline
            & \multicolumn{1}{l|}{Accuracy} & \multicolumn{1}{l|}{AUC-ROC Score} & F1 Score & \multicolumn{1}{l|}{Accuracy} & \multicolumn{1}{l|}{AUC-ROC Score} & F1 Score \\ \hline
GCN with CE & \multicolumn{1}{l|}{0.804}    & \multicolumn{1}{l|}{0.959}         & 0.783    & \multicolumn{1}{l|}{0.673}    & \multicolumn{1}{l|}{0.897}         & 0.609    \\ \hline
GCN with FL & \multicolumn{1}{l|}{0.802}    & \multicolumn{1}{l|}{0.971}         & 0.785    & \multicolumn{1}{l|}{0.648}    & \multicolumn{1}{l|}{0.875}         & 0.608    \\ \hline
GraphSMOTE  & \multicolumn{1}{l|}{0.774}    & \multicolumn{1}{l|}{0.953}         & 0.768    & \multicolumn{1}{l|}{0.31}     & \multicolumn{1}{l|}{0.632}         & 0.283    \\ \hline
Reg with CE & \multicolumn{1}{l|}{0.827}    & \multicolumn{1}{l|}{0.975}         & \textbf{0.806}    & \multicolumn{1}{l|}{0.683}    & \multicolumn{1}{l|}{0.895}         & \textbf{0.640}    \\ \hline
Reg with FL & \multicolumn{1}{l|}{0.823}    & \multicolumn{1}{l|}{0.976}         & 0.805    & \multicolumn{1}{l|}{0.668}    & \multicolumn{1}{l|}{0.893}         & 0.615    \\ \hline
\end{tabular}
\caption{Results for Cora and Citeseer Datasets}
\label{tab:results}
\end{table*}

Given a graph $G$ with $V$ nodes and $E$ edges, each node is labeled with one of the $N$ class labels. Further each node has a set of feature embeddings $f$. Let $m$ be the number of features and $X$ is a feature embedding matrix with shape $(V,m)$. The connectivity and structure of nodes is represented by the adjacency matrix $Adj$. 
Graph Attention networks have been used widely to do node classification \cite{velivckovic2017graph}. They do the node classification based on $X$ and attention mechanism $A$. 
The attention mechanism works as follows:
\begin{equation}
    h^{(k)}_{v} = f^{(k)}( W^{(k)}.[\sum_{u \in N(v)} a^{(k-1)}_{vu}h^{(k-1)}_{u} + a^{k}_{vv}h^{(k-1)}_{v}])
\end{equation}
Here $ h^{(k)}_{v}$ is embeddings of node $v$ at step $k$, where $h^{(0)}_{v}$ represents input feature embeddings of the node $x_{v} \in X$. This is calculated for all nodes $V$ in the graph. The attention weights $a^{k}$  are generated by an attention mechanism $A^{k}$ as follows  normalized such that the sum over all neighbours of each node $v$ is 1:
\begin{equation}
    a_{vu}^{(k)} = \frac{A^{(k)}(h_{v}^{(k)},h_{u}^{(k)})}{\sum_{w \in N(v)}(A^{(k)}(h_{v}^{(k)},h_{w}^{(k)}))}   (v,u) \in E
\end{equation}
Multiple attention heads are used instead of one head as it helps to train the model better. The output of the hidden layers from multiple attention heads is concatenated to get the input of the next hidden layer. We obtain a probability distribution of the $N$ classes based on the output of the last hidden layer.
 This model is trained by using the cross entropy loss function ($CEloss$).
\begin{equation}
    \hat{y_{v}} = F(h^{(K)}_{v})
\end{equation}
$\hat{y_{v}}$ is the prediction of node $v$. $K$ is the last hidden layer.
\par
In its present form, the model does not work well for an imbalanced dataset. Further investigation reveals that the model focuses more on the majority nodes. We propose a regularizer to make the model focus on minority nodes. We modify the loss function as follows:
\begin{equation}
    loss_{train}= loss + \lambda*(regularization) 
\end{equation}
$loss_{train}$ is the loss function used for training the model. $loss$ is a base loss function that is to be used with the regularizer. $regularization$ is a regularizer used to handle the imbalance. $\lambda$ is a hyperparameter used whose value can be between 0 to 1.
\subsection{Loss functions used}
We use two types of base loss functions which are used specifically when the data is imbalanced. Weighted cross entropy loss and Focal loss \cite{lin2017focal}. Weighted cross-entropy loss assigns every class different weights based on the number of samples in the class. The weights are assigned as follows:
\begin{equation}
    W_{c}=\frac{1}{\sqrt{N(c)}}
\end{equation}
$N(c)$ is the number of samples in class $c$. This ensures that classes with less samples that is, the minority classes get more weightage than the majority class while calculating the loss.\par
\citet{lin2017focal} proposed an algorithm that was helping solve the problem of 
extreme class imbalance in object detection problems by suggesting Focal loss. We use focal loss in the context of graphs. Focal loss reshapes the cross-entropy 
loss in such a way that it reduces the impact of easily classified examples and stresses more 
on the hard classified examples. This is done by multiplying the cross-entropy loss by a 
modulating factor, $(1 - pt)^\gamma$. The hyperparameter $\gamma$ acts as a scaling factor. It adjusts the rate at 
which easy examples are down-weighted. For easily classified examples, where pt tends 
towards 1, it causes the modulating factor to approach 0 and helps in reducing the sample’s 
effect on the loss. Focal loss is calculated according to the following equation.
\begin{equation}
    CE(p_t)=log(p_t)
\end{equation}
\begin{equation}
    FL(p_t)={-(1-p_t)}^{\gamma}log(p_t)
\end{equation}
As the value of $\gamma$ increases the model focuses more on the misclassified examples. Experiments have shown that using 0.6 as the $\gamma$ value gives optimal results.
\par
\subsection{Regularization}
We propose the following regularizer $KLDiv_{att}$. Out of all the attention heads used, we make one of the attention heads focus solely on the minority nodes. We find out $Adj_{minority} \subset Adj$ by taking rows of only the minority nodes from the adjacency matrix. Similarly, we find out $a_{minority} \subset a$ by taking rows of only minority nodes from the attention weights. We calculate the attention regularizer term by taking KLDivergence of $Adj_{minority}$ and $a_{minority}$. KLDivergence is used to measure the difference between two distributions. The $Adj_{minority}$ matrix consists of zeros and ones. Its value is one when the node has a minority node as its neighbor and is zero otherwise. $a_{minority}$ consists of the weights of minority nodes which are between zero and one. We want to maximise the weights and bring them numerically close to one of those nodes whose neighbours are minority nodes. This is similar to the distribution in $Adj_{minority}$. Therefore, we calculate KLDivergence as follows:

\begin{equation}
    KLDiv_{att}= Adj_{minority}*log(\frac{Adj_{minority}}{a_{minority}})
\end{equation}
We suggest a new loss function by adding $ KLDiv_{att}$ to the cross entropy loss function already used.
\begin{equation}
    loss_{train}= loss + \lambda*(KLDiv_{att}) 
\end{equation}
The model is then trained on this custom loss function. $\lambda$ is a hyperparameter which can be tuned between 0 to 1. Both the loss functions that are used aid in handling the imbalance along with the regularizer that we are using.\par

\section{Experiments}
\subsection{Datasets}
The datasets used in these experiments are Cora and CiteSeer. These are widely used benchmark datasets \cite{cora}. Both are citation network datasets. The number of edges, nodes, node features and the number of classes are shown in Table \ref{tab:description}.
\begin{table}[ht]
\centering
\begin{tabular}{|l|l|l|}
\hline
                   & Cora & Citeseer \\ \hline
Number of Nodes    & 2708 & 3327     \\ \hline
Number of Edges    & 5429 & 4732     \\ \hline
Number of Features & 1433 & 3703     \\ \hline
Number of classes  & 7    & 6        \\ \hline
\end{tabular}
\caption{Description the dataset used}
\label{tab:description}

\end{table}
The distribution of classes is shown in Table \ref{tab:dis}. We observe that the distribution of classes is uneven and imbalanced. We assume L1, L3, L5 and L6 as minority classes for Cora and the rest as majority classes. Similarly, we assume L3 and L4 as our minority classes in Citeseer.
\begin{table}[ht]
\centering
\begin{tabular}{|l|l|l|l|l|l|l|l|}
\hline
\textbf{Labels} & L0 & L1 & L2 & L3 & L4 & L5 & L6 \\ \hline
Cora         & 29 & 9*  & 16 & 13* & 15 & 11* & 7*  \\ \hline
Citeseer     & 18 & 20 & 21 & 8*  & 15* & 18 & -  \\ \hline
\end{tabular}
\caption{Distribution of classes (\%). (*=minority class)}
\label{tab:dis}
\end{table}

\textbf{Our experimental settings are as follows}: A semi-supervised setup is used for node classification. 
Each class has 20 examples. The validation set has 500 nodes, and the test set has 1000 nodes. Learning rate is set to 5e-3 and weight decay is set to 5e-4 for all experiments. The test set has the same minority nodes as the training set as used by standard graph methods. We use a total of two attention heads while using Graph Attention networks.
We vary the number of epochs varies between 100-1000. We found that training the model for 300 epochs had the best results while using the regularizer. We report the macro F1 score and the individual F1 scores of the minority nodes.

Table \ref{tab:results} shows the results of node classification for Cora and CiteSeer datasets. The methods that we use are: GCNs with weighted CE, GCNs with FL, existing GraphSMOTE method, Regularizer with weighted CE and Regularizer with FL.
The metrics that we report are accuracy, AUC-ROC score and the macro F1 score. The macro-F1 score is the most suitable and important metric for node classification.

\par
\begin{table}[ht]
\centering
\begin{tabular}{|l|llll|}
\hline
\textbf{Methods}     & \multicolumn{4}{l|}{\textbf{Cora minority classes}}                                              \\ \hline
            & \multicolumn{1}{l|}{L1}   & \multicolumn{1}{l|}{L3}   & \multicolumn{1}{l|}{L5}   & L6   \\ \hline
GCN with CE & \multicolumn{1}{l|}{0.65} & \multicolumn{1}{l|}{0.78} & \multicolumn{1}{l|}{0.77} & 0.71 \\ \hline
GCN with FL & \multicolumn{1}{l|}{0.85} & \multicolumn{1}{l|}{0.88} & \multicolumn{1}{l|}{0.74} & 0.73 \\ \hline
GraphSMOTE  & \multicolumn{1}{l|}{0.67} & \multicolumn{1}{l|}{0.88} & \multicolumn{1}{l|}{0.78} & 0.64 \\ \hline
Reg with CE & \multicolumn{1}{l|}{{0.85}} & \multicolumn{1}{l|}{0.92} & \multicolumn{1}{l|}{0.74} & 0.73 \\ \hline
Reg with FL & \multicolumn{1}{l|}{\textbf{0.85}} & \multicolumn{1}{l|}{\textbf{0.93}} & \multicolumn{1}{l|}{\textbf{0.74}} & \textbf{0.73} \\ \hline
\end{tabular}
\caption{Individual F1 scores for minority classes of Cora Dataset}
\label{tab:cora}
\end{table}


\begin{table}[ht]
\centering
\begin{tabular}{|l|ll|}
\hline
\textbf{Methods}     & \multicolumn{2}{l|}{\begin{tabular}[c]{@{}l@{}}\textbf{CiteSeer}\\  \textbf{minority classes}\end{tabular}} \\ \hline
            & \multicolumn{1}{l|}{L3}                                & L4                               \\ \hline
GCN with CE & \multicolumn{1}{l|}{0.24}                              & 0.66                             \\ \hline
GCN with FL & \multicolumn{1}{l|}{0.29}                              & 0.73                             \\ \hline
GraphSMOTE  & \multicolumn{1}{l|}{0.17}                              & 0.14                             \\ \hline
Reg with CE & \multicolumn{1}{l|}{0.32}                              & 0.71                             \\ \hline
Reg with FL & \multicolumn{1}{l|}{\textbf{0.26}}                              & \textbf{0.71}                             \\ \hline
\end{tabular}
\caption{Individual F1 scores for minority classes of CiteSeer Dataset}
\label{tab:citeseer}
\end{table}

\textbf{Comparison with GCNs}: We run the experiments on Graph Convolutional Networks as our baseline using both weighted CE and FL as our loss. We see that all three performance metrics: Accuracy, AUC-ROC Score, and F1 Score, increase with the addition of Regularizer as shown in table \ref{tab:results}. The individual F1 scores of minority nodes also increase, as shown in Table \ref{tab:cora} for Cora and in Table \ref{tab:citeseer} for CiteSeer dataset. \par
\textbf{Comparison with GraphSMOTE}: We compare our results with the existing state-of-the-art method to handle imbalance in Graph networks. We see a considerable rise in the F1 scores for both Cora and CiteSeer datasets, as shown in Table \ref{tab:results}. \par
 The method $Reg+CE$ works the best as it has the highest F1 Score. There is a rise in the individual minority F1 scores also apart from L5, as shown in Table \ref{tab:cora} for the Cora dataset.
The rise in the individual F1 Scores in minority classes, as shown in Tables \ref{tab:cora}, \ref{tab:citeseer} indicates that the algorithm is focusing more on the minority nodes than before.
\section{Conclusion and Future work}
In this work, we try to solve the problem of node imbalance by making intrinsic changes to the graph attention model. We do this by using Graph Attention networks, making the weights focus more on the minority nodes than the majority nodes. We use a custom loss function to train the model with these constraints. Concretely, we do not change the data composition in any way because the change in data can lead to information loss or have unnecessary noise in the model. 
 We observe that altering the attention weights by the addition of a regularizer significantly improves the results. We report the improvements in Cora and CiteSeer datasets compared to the existing methods. Individual F1 scores of minority classes empirically show that our method helps mitigate the imbalance by focusing more on minority classes.\par
We plan to implement similar techniques on larger datasets. We aim to check if this Regularization-based method or other intrinsic methods work for other tasks like edge-detection and graph detection. We also intend to explore how the sparsity of graphs is related to handling the imbalance.
\bibliography{aaai23.bib}
\end{document}